\documentclass{article}

\usepackage{arxiv}

\usepackage[utf8]{inputenc} 
\usepackage[T1]{fontenc}    
\usepackage{hyperref}       
\usepackage{url}            
\usepackage{booktabs}       
\usepackage{amsfonts}       
\usepackage{nicefrac}       
\usepackage{microtype}      
\usepackage{lipsum}

\usepackage{graphicx}
\usepackage{algorithmic}

\algsetup{linenosize=\small}
\usepackage[ruled,vlined]{algorithm2e}
\SetAlFnt{\small}
\SetAlCapFnt{\small}
\SetAlCapNameFnt{\small}

\usepackage{caption}
\usepackage{subcaption}
\usepackage{mwe}
\usepackage{comment}
\usepackage{enumitem}

\title{Generation of Synthetic Electronic Health Records Using a Federated GAN}

\author{
   John Weldon\\
   School of Computing\\
   Dublin City University\\
   Ireland\\
   \texttt{john.weldon8@mail.dcu.ie} \\
   \And
  Tom\'{a}s Ward \\
   Insight SFI Centre for Data Analytics\\
   Dublin City University\\
   Ireland\\
   \And
  Eoin Brophy \\
   School of Computing and INFANT Research Centre\\
   Dublin City University\\
   Ireland\\
 }

\begin{document}
\maketitle

\begin{abstract}
Sensitive medical data is often subject to strict usage constraints. In this paper, we trained a generative adversarial network (GAN) on real-world electronic health records (EHR). It was then used to create a data-set of "fake" patients through synthetic data generation (SDG) to circumvent usage constraints. This real-world data was tabular, binary, intensive care unit (ICU) patient diagnosis data. The entire data-set was split into separate data silos to mimic real-world scenarios where multiple ICU units across different hospitals may have similarly structured data-sets within their own organisations but do not have access to each other's data-sets. We implemented federated learning (FL) to train separate GANs locally at each organisation, using their unique data silo and then combining the GANs into a single central GAN, without any siloed data ever being exposed. This global, central GAN was then used to generate the synthetic patients data-set. We performed an evaluation of these synthetic patients with statistical measures and through a structured review by a group of medical professionals. It was shown that there was no significant reduction in the quality of the synthetic EHR when we moved between training a single central model and training on separate data silos with individual models before combining them into a central model. This was true for both the statistical evaluation (Root Mean Square Error (RMSE) of 0.0154 for single-source vs. RMSE of 0.0169 for dual-source federated) and also for the medical professionals' evaluation (no quality difference between EHR generated from a single source and EHR generated from multiple sources).
\end{abstract}


\section{Introduction}

Real data sets often have strict usage constraints due to privacy rules and regulations. Generated synthetic versions of these data sets can replicate the statistical qualities of actual data without exposing the data, eliminating these constraints. This is critical when dealing with human subjects' data, protecting these subjects' privacy and confidentiality. There are other motivations behind SDG besides usage constraints. Typically when training machine learning algorithms, the more training data, the better the results and the better the trained models will perform. Often there is a lack of data, or the data is expensive to generate. For example, in the autonomous vehicle space, producing synthetic data is far more cost-effective and efficient than collecting real-world data \cite{Yurtsever_2020}. One can also use synthetic data to test existing models; instead of using costly real-world data to test.

FL was first introduced by Google in their 2016 paper \cite{mcmahan2017communicationefficient}. While their vision involved mobile devices being the decentralised clients in which training occurs before the models are aggregated in a central location, their techniques are equally suited to larger entities such as financial institutions and hospitals. FL allows institutions to share the value of their private data with other similar institutions without sharing actual data and in return receive the value of these other institutions' data. While it is possible to use a peer to peer structure that does not require a centralised model as seen in \cite{roy2019braintorrent}, we wanted to examine the centralised implementation.

\subsection{PhysioNet's MIMIC-III Electronic Health Records}
\begin{figure}[ht]
\begin{center}
\includegraphics[width=0.65\columnwidth]{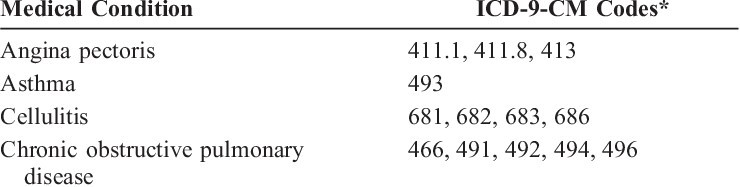}
\caption{Example of ICD9 Diagnosis Codes \cite{p} } \label{fig:ICD_codes}
\end{center}
\end{figure}

The EHR that we used to create synthetic data was PhysioNet's \cite{gaghim00} MIMIC-III 1.4 \cite{jpm16}. This data is an example of the strict usage constraints mentioned previously. We had to complete a human research ethics course and wait 5 weeks to gain access to it. It's a sizeable relational database detailing patients admitted to ICU at Beth Israel Deaconess Medical in Boston, MA, USA. The data we wanted to synthesise was patient ICD-9 diagnoses. Figure~\ref{fig:ICD_codes} shows the structure with a sample list from the 1071 unique codes in the data set.
f

\subsection{Generative Adversarial Networks}
GANs typically consist of two separate neural networks, the generator (G) and the discriminator (D, sometimes called the critic). G takes in Gaussian random noise (Z) and attempts to generate synthetic data similar to the trained data. D attempts to determine whether or not the data produced by G is fake or if it's data taken from the real data-set. G attempts to maximise how often D classifies the fake data as being real and D simultaneously attempting to maximise how often it correctly classifies it as fake. An often-used analogy is that of counterfeiting. G, in this analogy, is a criminal who is creating counterfeit banknotes, and D represents the bank teller attempting to detect if the notes are fraudulent. As the criminal becomes better, producing notes that are more and more realistic, the teller must also improve in ability to distinguish these counterfeit notes from the real thing. The reverse is also true; as the teller improves at detecting the counterfeit notes, the criminal must increase the quality if they want to succeed in fooling them.  Figure~\ref{fig:gan_network} below shows how the GAN is structured. \cite{goodfellow2014generative}.

\begin{figure}[ht]
\begin{center}
\includegraphics[width=.65\textwidth]{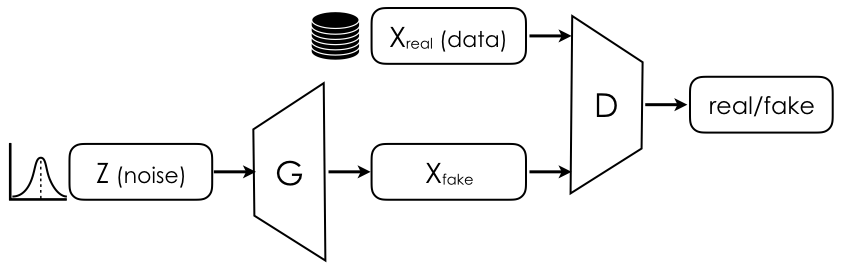}
\caption{GAN Architecture Overview} \label{fig:gan_network}
\end{center}
\end{figure}

\section{Related Work}
Highly relevant is medGAN; Choi et al. demonstrated that it was possible to use a GAN they created to generate high-quality synthetic EHR \cite{choi2018generating}. Rasouli et al. showed that it was possible to use FL techniques with GANs to provide decentralised training of both G and D \cite{rasouli2020fedgan}. Torfi and Fox used medical professionals to evaluate the data quality of generated EHR though it is not clear how this evaluation was structured \cite{corgan}. Finally, Brophy et al. demonstrated that a federated GAN could be successfully used to generate continuous time-series EHR. \cite{brophy2021estimation}. 

As far as we are aware, the work in this paper is novel in that it is the first attempt to use a federated GAN for the generation of discrete, binary EHR. It also contains a novel evaluation methodology, providing a repeatable method for medical professionals to evaluate these synthetic EHR.

\section{Implementation}

\subsection{Data Pre-processing}
\begin{figure}[ht]
\begin{center}
\includegraphics[width=.85\textwidth]{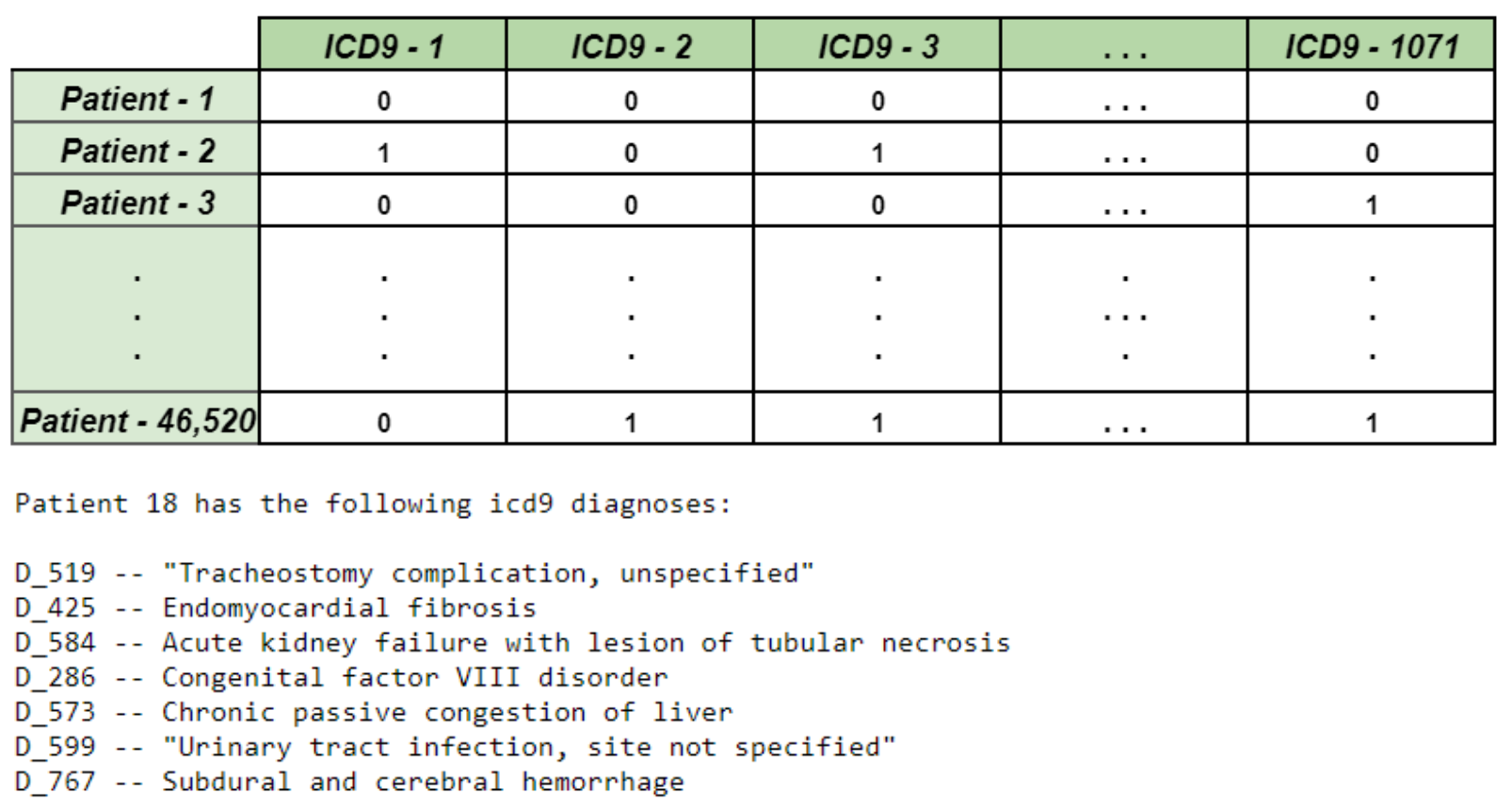}
\caption{Feature vectors and sample patient} \label{fig:features}
\end{center}
\end{figure}

To process the MIMIC-III data-set into the required format, we followed the process documented in \cite{choi2018generating}. This left us with the data structure in Table~\ref{table:patient_data}. The feature columns contain binary data, with 1 indicating that the patient received that ICD-9 diagnosis and 0 indicating they did not. Figure~\ref{fig:features} shows what this looked like and how the patients were presented once the ICD-9 codes were mapped to descriptive diagnoses.

\begin{table}[ht]
\begin{center}
\setlength{\tabcolsep}{8pt}
\caption{Table showing structure of patient data}\label{table:patient_data}
\begin{tabular}{|l|l|l|}
\hline
Data-set Version Info &  ICU Patients (Rows) & ICD9 Diagnoses (Columns)\\
\hline
MIMIC-III 1.4 & 46,520 & 1,071\\
\hline
\end{tabular}
\end{center}
\end{table}

\subsection{GAN architecture}
The GAN was created with Python 3.8, using Tensorflow-GPU 2.2. Both G and D had depths of 3, with the dimension sizes being the reverse of each other. As this was tabular data and not images, convolutional layers were unnecessary and dense layers were used in their place. The activation function for the first three layers in both G and D were Leaky Rectified Linear Units (Leaky RELU). The final layer in D had a sigmoid activation function with range [0,1] and midpoint at 0.5, and the last layer in G had a tanh activation function with range [-1,1] and centre at 0. Table~\ref{table:GAN} shows the details of both G and D.

\subsection{GAN training}
The hyperparameters used within the GAN were based on choices made by Goodfellow et al. \cite{goodfellow2014generative}. We experimented with many different combinations of epochs, batch size, learning rate, optimiser. We found that for our use case, the following hyperparameters gave us the best results.

\textbf{Learning Rate:} 0.0002, \textbf{Batch Size:} 1,500, \textbf{Epochs:} 20,000 total (10,000 per local model for 2 source, 4,000 per local model for 5 source etc.), \textbf{Optimiser:} ADAM.

For each weight update within G, we made two consecutive updates within D, a method that has been known to increase the quality of the synthetic data in some cases and proved to be true in our case also \cite{ganhack}.

\begin{table}[ht]
\centering
\setlength{\tabcolsep}{4pt}
\caption{GAN Architecture}\label{table:GAN}
\begin{tabular}{|l|l|l|}
\hline
GAN Component & Layer &  Dimension (G)\\
\hline
Discriminator (D) & Input &  1,071\\
Discriminator (D) & 1 - Leaky RELU & Dense 512\\
Discriminator (D) & 2 - Leaky RELU & Dense 256 \\
Discriminator (D) & 3 - Leaky RELU & Dense 128\\
Discriminator (D) & Output & Dense 1, Sigmoid activation\\
Generator (G) & Input &  Gaussian Random Noise 128\\
Generator (G) & 1 - Leaky RELU & Dense 128\\
Generator (G) & 2 - Leaky RELU & Dense 256 \\
Generator (G) & 3 - Leaky RELU & Dense 512\\
Generator (G) & Output & Dense 1,071, Tanh activation\\
\hline
\end{tabular}
\end{table}

\subsection{Federated Learning Structure}

\begin{figure}[ht]
\begin{center}
\includegraphics[width=0.65\columnwidth]{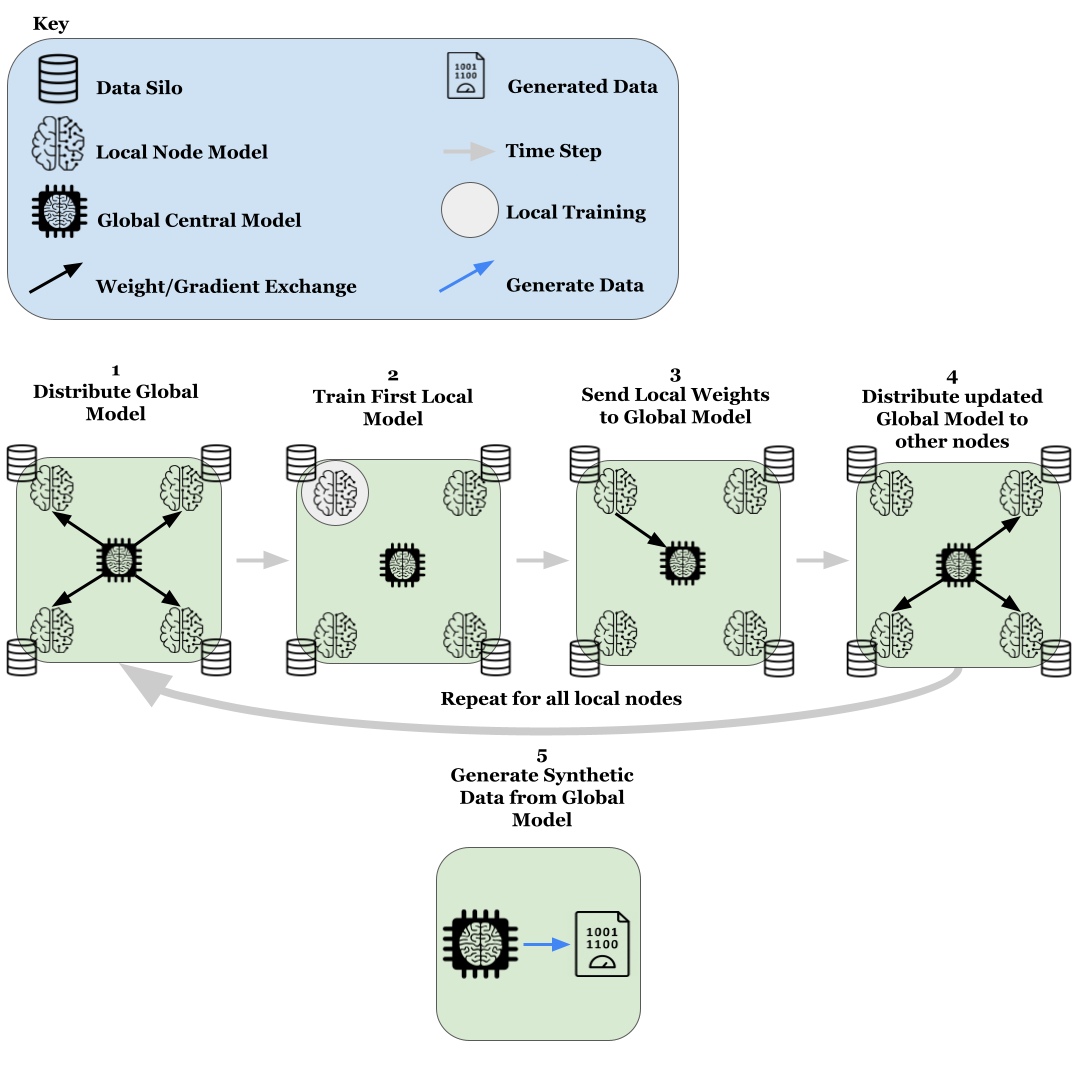}
\caption{Centralised Federated Learning Process} \label{fig:FedLearning}
\end{center}
\end{figure}

Algorithm~\ref{alg:1} and Figure~\ref{fig:FedLearning} show the structure for a single round of our FL implementation. For actual use cases, these rounds may occur at some periodic interval to ensure each client has the most up-to-date model or maybe a new round is initiated if one of the local clients has a significant increase in data. Thus, the process is flexible and could be adjusted to whatever is most suited to the specific use case. What's important is that all local clients will have the latest version of the global model by the end of one full round.

\begin{algorithm}[ht]
\SetAlgoLined
 create global GAN\;
 create local node GANs\;
 create nodelist containing all local nodes\;
 
 \For{local node in nodelist}{
  retrieve weights from global GAN\;
  set local node GAN weights to global weights\;
  train local node GAN on local data silo\;
  set global GAN weights to local node GAN weights\;
 }
 
 use final global GAN to generate synthetic patients\;
 \caption{Federated Learning Process Pseudo-code}\label{alg:1}
\end{algorithm}

\section{Evaluation}
\subsection{Synthetic Data Quality}
\subsubsection{Feature Probability Correlation - Visual Comparison:}
\begin{figure*}[ht]
    \centering
    \begin{subfigure}[b]{0.45\textwidth}
        \centering
        \includegraphics[width=.99\textwidth]{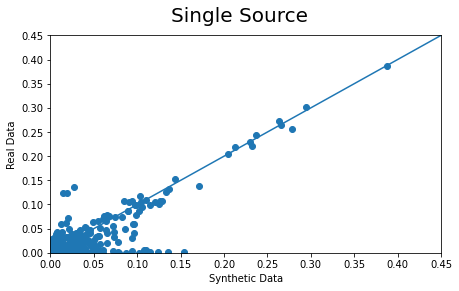}
        \caption[Network2]%
        {{\small Full 46,520}}    
        \label{fig:mean and std of net14}
    \end{subfigure}
    \hfill
    \begin{subfigure}[b]{0.45\textwidth}  
        \centering 
        \includegraphics[width=.99\textwidth]{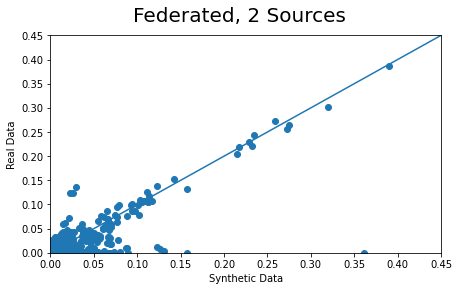}
        \caption[]%
        {{\small Two silos of 23,260}}    
        \label{fig:mean and std of net24}
    \end{subfigure}
    \vskip\baselineskip
    \begin{subfigure}[b]{0.45\textwidth}   
        \centering 
        \includegraphics[width=.99\textwidth]{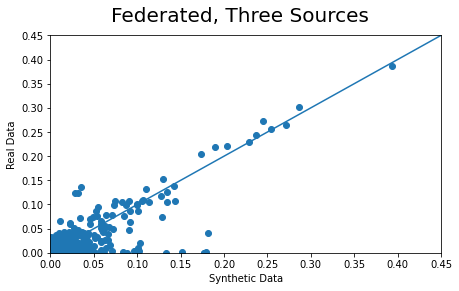}
        \caption[]%
        {{\small Three silos of 15,506}}    
        \label{fig:mean and std of net34}
    \end{subfigure}
    \hfill
    \begin{subfigure}[b]{0.45\textwidth}   
        \centering 
        \includegraphics[width=.99\textwidth]{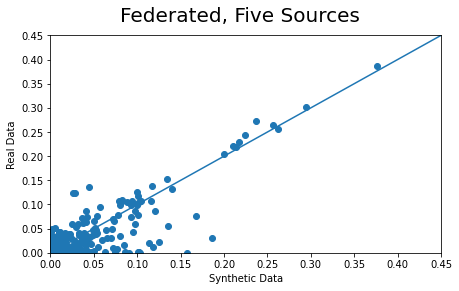}
        \caption[]%
        {{\small Five silos of 9,304}}    
        \label{fig:mean and std of net44}
    \end{subfigure}
    \caption[ Feature probability comparison between single and federated models ]
    {\small Feature probability comparison between single and federated models } 
    \label{fig:mean and std of nets}
\end{figure*}

The first step that we took when evaluating the quality of the synthetic patients was to visually compare how close they were to the actual patients regarding the probability of diagnoses. We calculated the probability that each of the 1,071 diagnosis features was present for any given patient in both the real and synthetic data and plotted these probabilities against each other. If they were perfectly matched, all 1,071 points would land perfectly on the X=Y line. We can see in Figure~\ref{fig:mean and std of nets} that for all models, the points were clustered along this line. Importantly, we did not see any dramatic decrease in quality as we moved from the single source model to the federated versions. Comparing the single-source model to the dual-source federated model showed no loss in quality at this high level. We seemed to start to lose some quality when we moved on to three sources and five sources. While this was a helpful introduction to the evaluation process, we needed to make more concrete numerical comparisons.

\subsubsection{Feature Probability Correlation - Statistical Comparison:}
We used two statistical measures to compare the synthetic data sets from the various models to the real data in diagnosis probability—coefficient of Determination (R Squared) and Root Mean Square Error (RMSE). RMSE indicates the absolute fit of our points to the ideal line, in other words, how close the synthetic diagnosis probabilities are to the actual diagnosis probabilities. Whereas R-squared is a relative measure of how closely correlated the two probability vectors are. Lower values of RMSE indicate a better fit, whereas higher values of R Squared indicate the two sets of probabilities are more highly correlated. We can see in Table~\ref{table:CC} that while RMSE starts to increase and R Squared starts to decrease, as we split the data-set into more and more silos, the changes are negligible, and we did not deem them significant.

\begin{table}[ht]
\begin{center}
\setlength{\tabcolsep}{8pt}
\caption{Feature Probability Correlation Comparison}\label{table:CC}
\begin{tabular}{|l|l|l|}
\hline
Model & R Squared & RMSE\\
\hline
Single & 0.805 & 0.0154\\
Federated - 2 Silos & 0.793 & 0.0161\\
Federated - 3 Silos & 0.786 & 0.0169\\
Federated - 5 Silos & 0.782 & 0.0174\\
\hline
\end{tabular}
\end{center}
\end{table}

\subsubsection{Medical Professional Evaluations:}
Many GAN implementations involve image data such as digits (MNIST) \cite{jain2020locally}, or cat images \cite{huang2018multimodal}. While it is straightforward to derive a feel for the quality of the images in these cases simply by glancing at them, it's not so simple for someone who lacks the specific domain knowledge to do the same for EHR. While the previous metrics suggested our synthetic data was similar to the real data, we wanted to seek input from experts in this field. A group of Medical Professionals was tasked with evaluating the generated patients. This group consisted of the following individuals:

\begin{itemize}
\item A Critical Care Doctor within the ICU unit in Beaumont Hospital, Dublin, Ireland,
\item A Critical Care Nurse within the ICU unit in the Mater Private Hospital, Dublin, Ireland,
\item An Anatomical Pathologist from the Ottawa Hospital, Ontario, Canada.
\end{itemize}

The evaluation contained 20 actual patients, 20 patients generated by the single source GAN and 20 patients generated by the dual-source, federated GAN. The patients were randomly sorted and unlabelled. The guidelines were as follows:

\textit{"Please rate these 60 patients in terms of how plausible you consider them to be as ICU patients that
you might encounter on a given day in your ICU or any ICU. Please try not to take into account how likely or unlikely it may be to encounter one of these patients, as rarity should not be a factor. 
\\
* Please note that this list is a mixture of real patients (from an ICU in the United States) and generated synthetic patients."}

They were asked to place each patient into one of six categories. We chose an even number so there was no middle category that would allow them to sit on the fence when evaluating any patient. These were the six categories:

\begin{itemize}[leftmargin=1.2in]
\item \textit{Highly Plausible,}
\item \textit{Plausible,}
\item \textit{Slightly Plausible,}
\item \textit{Slightly Implausible,}
\item \textit{Implausible,}
\item \textit{Highly Implausible.}
\end{itemize}

\begin{figure}[ht]
\begin{center}
\includegraphics[width=.7\textwidth]{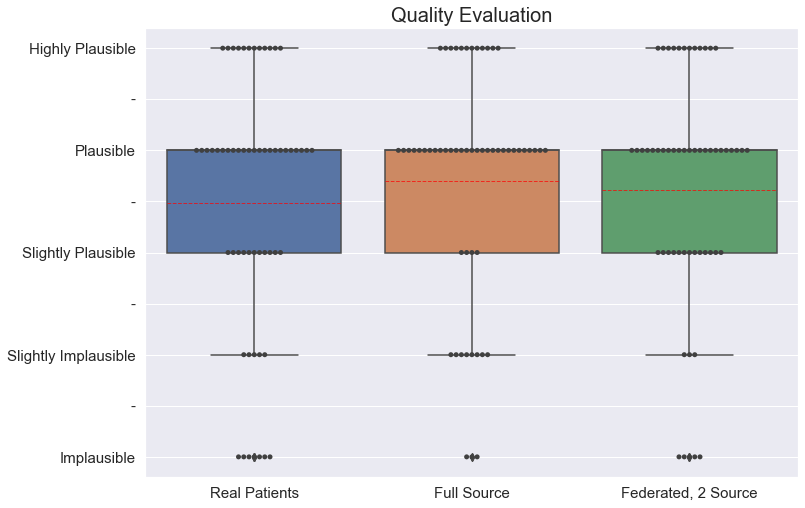}
\caption{Medical Professionals' Evaluation} \label{fig:Evaluation}
\end{center}
\end{figure}

Figure~\ref{fig:Evaluation} shows us that there was no significant difference between the three groups of patients regarding how plausible the medical professionals viewed them, confirming that the synthetic patients closely matched the real patients for both the single source GAN and the federated GAN with two separate sources. Unfortunately, it was not possible to include the three-source GAN and the five-source GAN as these professionals are extremely busy, and though they were all generous with their time, they could only dedicate so much.

\subsection{Data Privacy}
We implemented two different checks for patient confidentiality, detailed below. It's important to note that while these checks provide some reassurance that the generated data is private and confidential, neither check can sufficiently verify that an actual patient could not be identified via a well-orchestrated membership inference attack on the synthetic data by some malevolent actor \cite{hu2021membership}.

\subsubsection{Checking for exact matches:}
The first privacy measure to determine if any actual patient data was exposed was a duplicate check. This involved iterating across the 46,520 real patients and see if their exact diagnoses vector had a perfect match in the 46,520 synthetic patients. If so, it would mean that any potential attacker who knew a patient's diagnoses could quickly tell if this patient was a part of the data set. We can see in Table~\ref{table:Privacy} that there are no exact matches between the synthetic patients and our real patients for all four models. Though this is a reasonably simplistic measure, it serves as a quick sanity check and gives some confidence that privacy has been attained.

\subsubsection{Similarity Threshold:}
\begin{figure*}[ht]
    \centering
    \begin{subfigure}[b]{0.475\textwidth}
        \centering
        \includegraphics[width=.75\textwidth]{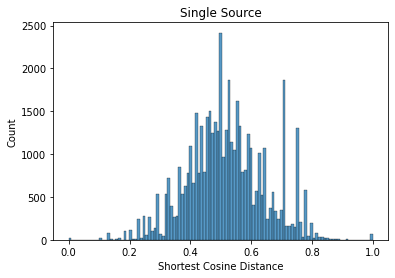}
        \caption[Network2]%
        {{\small Full 46,520}}    
        \label{fig:cosine_net14}
    \end{subfigure}
    \hfill
    \begin{subfigure}[b]{0.475\textwidth}  
        \centering 
        \includegraphics[width=.75\textwidth]{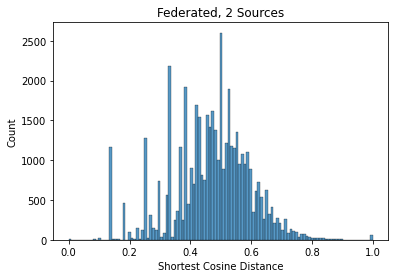}
        \caption[]%
        {{\small Two silos of 23,260}}    
        \label{fig:cosine_net24}
    \end{subfigure}
    \vskip\baselineskip
    \begin{subfigure}[b]{0.475\textwidth}   
        \centering 
        \includegraphics[width=.75\textwidth]{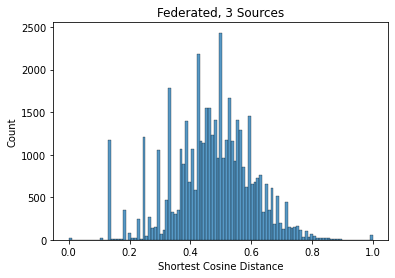}
        \caption[]%
        {{\small Three silos of 15,506}}    
        \label{fig:cosine_net34}
    \end{subfigure}
    \hfill
    \begin{subfigure}[b]{0.475\textwidth}   
        \centering 
        \includegraphics[width=.75\textwidth]{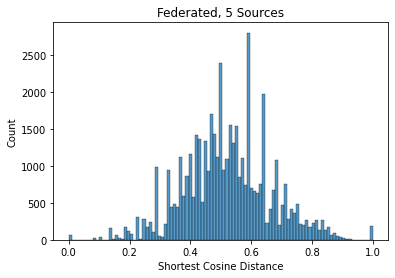}
        \caption[]%
        {{\small Five silos of 9,304}}    
        \label{fig:cosine_net44}
    \end{subfigure}
    \caption[ Distribution of Shortest Cosine Distances ]
    {\small Distribution of Shortest Cosine Distances } 
    \label{fig:Cosine_nets}
\end{figure*}

The second method was to find the cosine similarity between each real patient and all of the synthetic patients. While the synthetic data needs to be realistic, it was essential that none of the synthetic patients was so alike any real patient that it would be apparent they were part of the training data. Cosine similarity provides a more meaningful similarity metric between two diagnosis vectors than Euclidean distance does \cite{corgan}. This similarity is subtracted from 1 to derive distance so the lower the value, the more similar two patients are. Table~\ref{table:Privacy} shows the mean and standard deviation for shortest cosine distance for all patient pairs. Figure~\ref{fig:Cosine_nets} shows that the distribution of shortest cosine distances is approximately normal for all four. Very few patients fall within 0.1 of any real patients, and so the vast majority of synthetic patient pairs are below 0.9 for similarity, which we set as our threshold value.

\begin{table}[ht]
\setlength{\tabcolsep}{8pt}
\caption{Data Privacy}\label{table:Privacy}
\begin{center}
\begin{tabular}{|l|l|l|l|}
\hline
Model &  Duplicates & Mean Min Cos. Distance & Standard Dev.\\
\hline
Single    &  {\bfseries} 0 & 0.5194 & 0.1283 \\
Federated - 2 Silos    &  {\bfseries} 0 & 0.4698 & 0.1289 \\
Federated - 3 Silos    & {\bfseries} 0 & 0.4736 & 0.1334 \\
Federated - 5 Silos    & {\bfseries} 0 & 0.5229 & 0.1438 \\
\hline
\end{tabular}
\end{center}
\end{table}

\section{Discussion}

\subsection{Mode Collapse}

GANs sometimes show a lack of diversity amongst generated samples. This is known as Mode Collapse (MC) \cite{che2017mode}. We return to the counterfeiting analogy. Suppose the counterfeiter manages to create banknotes that consistently fool the teller. They will see no need to alter these notes and continue producing similar samples, knowing they're likely to pass as authentic. This represents a GAN suffering from MC. Suppose G stumbles upon a combination that consistently fools D. In that case, it will see no reason to diversify future samples and risk lowering the chance of fooling D. Though our GAN did produce diverse patients, there were signs of MC. In terms of the diagnosis combinations, certain patient types appeared often. One method for mitigating mode collapse is Wasserstein Loss (WGAN) \cite{arjovsky2017wasserstein}; specifically, a Wasserstein GAN with Gradient Penalty (WGAN-GP) \cite{gulrajani2017improved}. This enforces a penalty if G consistently produces highly similar samples, meaning that G cannot sit in this collapsed mode and must instead generate more diverse samples.

\subsubsection{Wasserstein GAN - Gradient Penalty:}
\begin{figure}[ht]
\begin{center}
\includegraphics[width=.45\textwidth]{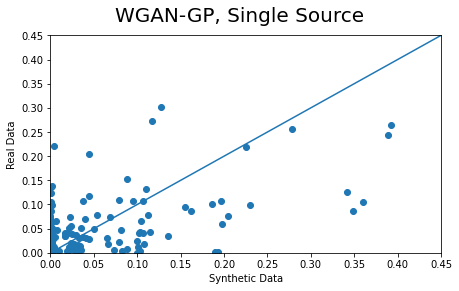}
\caption{WGAN-GP Feature Probability Comparison} \label{fig:ProbabilityComparison}
\end{center}
\end{figure}

We attempted to eliminate MC in our model by using WGAN-GP. Figure~\ref{fig:ProbabilityComparison} shows the quality of the WGAN-GP data. It did help mitigate MC, generating a more diverse range of patients, but these patients' quality suffered. The trade-off between MC and data quality has been documented by \cite{Adiga2018ONTT}. We decided that this quality drop was not acceptable and so we stuck with our previous model.

\section{Conclusion}

We propose that the federated GAN can adequately learn the distribution of real-world EHR and exhibit comparable performance to the single source GAN in generating realistic synthetic binary EHR. We analysed both the synthetic EHR data generated by the single source GAN and the synthetic EHR data generated by the federated GAN and compared their evaluation results with the actual EHR data. Based on our investigation, we conclude that there is no significant reduction in quality between the data generated by single-source GAN and the data generated by the federated GAN. A potential idea for future research could be to combine the federated GAN with more formal privacy-preserving mechanisms like differential privacy to provide stronger privacy guarantees. Another avenue to explore would be improving the WGAN-GP implementation to eliminate MC without significant data quality loss.

\bibliographystyle{acm}  
\bibliography{main} 

\end{document}